# A Light-Weight Object Detection Framework with FPA Module for Optical Remote Sensing Imagery

Xi Gu
Department of Computing Science, Tongji University, China
1833015@tongji.edu.cn

Lingbin Kong
Department of Software Engineering, Tsinghua University, China
klb18@mails.tsinghua.edu.cn

Zhicheng Wang *
Department of Computing Science, Tongji University, China
zhichengwang@tongji.edu.cn

Jie Li
Shanghai Aerospaces Electronic Technology Institute, Shanghai, China
trackerdsp@163.com

Zhaohui Yu
Shanghai Aerospaces Electronic Technology Institute, Shanghai, China

Gang Wei
Department of Computing Science, Tongji University, China

## ABSTRACT
With the development of remote sensing technology, the acquisition of remote sensing images is easier and easier, which provides sufficient data resources for the task of detecting remote sensing objects. However, how to detect objects quickly and accurately from many complex optical remote sensing images is a challenging hot issue. In this paper, we propose an efficient anchor free object detector, CenterFPANet. To pursue speed, we use a lightweight backbone and introduce the asymmetric revolution block. To improve the accuracy, we designed the FPA module, which links the feature maps of different levels, and introduces the attention mechanism to dynamically adjust the weights of each level of feature maps, which solves the problem of detection difficulty caused by large size range of remote sensing objects. This strategy can improve the accuracy of remote sensing image object detection without reducing the detection speed. On the DOTA dataset, CenterFPANet mAP is 64.00%, and FPS is 22.2, which is close to the accuracy of the anchor-based methods currently used and much faster than them. Compared with Faster RCNN, mAP is 6.76% lower but 60.87% faster. All in all, CenterFPANet achieves a balance between speed and accuracy in large-scale optical remote sensing object detection.

## CCS Concepts
•**Computing methodologies**➜**Artificial intelligence**➜**Computer vision**➜**Computer vision problems**➜**Object detection**

## 1. INTRODUCTION
Remote sensing image technology is increasingly used. For example, companies need to use remote sensing image technology to provide location-related services. Remote sensing images are used to build a complete and clear model of the earth and observe the movement of the earth's surface in real time. The government also uses remote sensing technology to provide weather forecasting, traffic forecasting and other services. With the increasing application of remote sensing technology, the datasets of remote sensing images are increasing[1]. The application of object detection in remote sensing images has become a hot issue.

In object detection, remote sensing image detection is more difficult than natural image. Object detection in large remote sensing images (such as ship, aircraft and vehicle detection) is a challenging task because of the small size, large number of objects, and complex surrounding environment, which may lead to the detection model mistaking unrelated ground objects as object objects. The objects in natural images are relatively large, and the limited scenes are not complex compared to remote sensing images, making it easier to identify objects. This is one of the main differences between detecting remote sensing images and natural images.

In recent years, many object detection methods based on deep learning have been proposed, which significantly improve the performance of object detection. Existing deep learning methods for object detection can be divided into two schools according to whether they are anchor-based or not. They are anchor-based regional proposals and anchor-free regression methods, respectively.

In the past few years, anchor-based object detection methods have achieved great success in natural scene images. This method divides the framework of object detection into two phases. The first phase focuses on generating a series of candidate regions that may contain objects. The second stage classifies the candidate regions obtained in the first stage as target objects or backgrounds, and further fine-tunes the coordinates of the bounding box. R-CNN[2] is one of the typical representative algorithms for object detection methods proposed by regions. R-CNN first obtains candidate regions with possible targets through selective search, then extracts the eigenvectors from the candidate regions using a convolution neural network, and then passes the eigenvectors into the support vector machine for classification. Although it introduces a neural network in the process of image feature extraction to avoid the drawbacks of artificial feature extraction, it is difficult to meet the speed requirements because of the convolution operation for each candidate region. Fast RCNN[3], Faster RCNN[4], Mask RCNN[5] and so on have made different improvements to R-CNN, resulting in excellent performance of the region proposal method.

Nevertheless, due to the limited storage space and computing power of mobile devices, anchor-based methods require tens of thousands of anchors to be preset in the image for a high recall rate, which makes it difficult to further improve its detection speed. Therefore, researchers began to try to optimize the anchor-free regression algorithm.

The anchor free regression algorithm uses a single-stage object detection model for prediction, simplifying the two steps of object detection. Regression-based methods are much simpler and more efficient than region-based methods because there is no need to generate candidate region proposals and subsequent feature resampling phases. The DenseBox[6] proposed in 2015 was an early work of anchor free regression algorithm. It uses full convolution network to achieve end-to-end training and recognition and has a good effect on face detection tasks. However, due to the late publication of this paper, anchor free algorithm enters people's view later than region proposal object detection method, so region proposal method occupies a dominant position in the field of object detection.

Redmon et al.[7] proposed YOLO (You Only Look Once), which uses object detection as a regression problem, divides the image into discrete grids, and directly returns the probability of bounding boxes and related classes. Since there is no region proposal generation stage, YOLO uses a set of candidate regions to directly predict results. YOLO is fast in design because the region proposal generation step has been completely abandoned. However, YOLO has a greater positioning error than RCNN because the bounding boxes are roughly divided.

Law et al.[8] questioned anchor's dominant role in the SOTA object detection framework. Law et al. argue that anchor has drawbacks such as causing a huge imbalance between positive and negative samples, slowing down training speed, and introducing additional hyperparameters. Law et al. borrowed the idea of Associative Embedding in multi-person posture estimation and considered the boundary box object detection as the key point to detect paired upper left and lower right corners. In CornerNet, the network consists of two stacked Hourglass networks, and a simple corner pooling method is used to better locate corners. Based on COCO, CornerNet achieves 42.1% AP, surpassing all previous single-stage detectors. However, CornerNet has difficulty deciding which key points should be grouped into the same target, which can lead to detection errors.

To further improve CornerNet, Duan et al.[9] proposed CenterNet. By introducing an additional key point in the center of the bounding box, corner matching is more accurate, the probability of mismatching corners in CornerNet is reduced, and COCO AP is increased to 47.0%, but this also results in slower matching than CornerNet.

When a large amount of computation cost is allowed, anchor-based region proposal algorithms usually achieve higher detection accuracy than anchor-free. Most of the winners in the well-known detection challenge are anchor-based.

However, although CNN has achieved great success in natural scene images, it is problematic to use it directly for object detection in optical remote sensing images because of the difficulty in effectively handling the rotation changes of objects. Essentially, this is not a critical issue for natural scene images, because objects are usually in a vertical direction due to the gravity of the earth, so the direction changes on the images are usually small. Conversely, objects in remote sensing images, such as airfields, buildings, and vehicles, often have many different directions because remote sensing images are taken from airspace over the sky. In addition, small objects account for a large proportion in remote sensing images, are densely distributed, and the conditions for taking images are unstable, which are easily affected by factors such as lighting, clouds, camouflage and so on. Therefore, the accuracy of object detection in remote sensing images is often lower than that in natural images.

Detection speed is also important in this area. For example, Sentinel-1 imagery the entire earth every six days. Even though Sentinel satellites were launched in 2014, they already have about 25 PB of data. The Coperni-cus concept requires that the algorithm be fast enough and transferable enough to be applied across the earth's surface[10].

However, most of the current remote sensing object detection algorithms are based on anchors, which has a lot of redundancy and high computational complexity, and the detection speed is relatively slow. The object detection model based on anchor free has lost some precision, but it often has faster detection speed. Therefore, it is of great significance to further explore the object detection method based on anchor free for improving the application ability of detect remote sensing image.

The main contributions of this paper lie in the following two aspects.

(1) We present a simple multi-classification object detection framework, which can be applied to remote sensing dataset.

(2) We guarantee the accuracy of detecting the objects in the remote sensing image and make the detection speed greatly improved compared with the existing detector.

Compared with the first stage network YOLOv2 and YOLOv3, our network speed is almost the same, and mAP is greatly improved. Compared with the fast RCNN of the two-stage network, the mAP is slightly reduced, but the speed is twice. All in all, the proposed model achieves the balance of speed and map to achieve high detection performance in large-scale optical remote sensing images.

## 2. Methods

This section mainly introduces the overall structure and details of the proposed network. Section 3.1 describes the overall network structure. Section 3.2 describes the encoder part of the network. Section 3.3 introduces the application of feature pyramid network. Section 3.4 introduces the structure of attention mechanism. Section 3.5 introduces the decoder part of the network and the inference process. Section 3.6 describes the loss function used in training.

### 2.1 Overall network structure

Our CenterFPANet is based on CenterNet, as shown in Figure 1. CenterFPANet is an anchor free network with encoder decoder structure. First of all, the encoder network mainly extracts the low-level and high-level features from the bottom-up, top-down and horizontal connection paths through the feature pyramid network, so that the model can better capture the global and local information. Then, the acquired features enable important channels to be assigned greater weight through channel attention mechanism. Finally, the feature image is transferred to the decoder network. The decoder part outputs heatmaps corresponding to each class, through which the probability of the center point of the object, the offset of the center point and the length and width of the center point object are obtained.

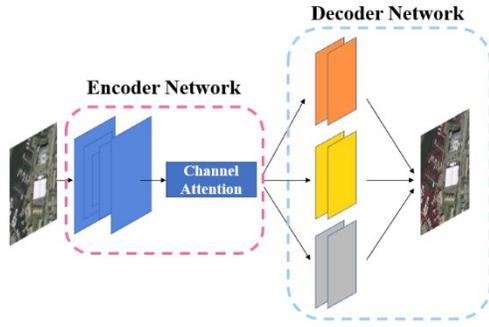

**Figure 1. The overall structure of CenterFPANet. Encoder network includes feature pyramid network and channel attention mechanism. Decoder network includes three network branches that predict object center, center offset and object size.**

## 2.2 Backbone network

Resnet[11] proposed in 2015 is widely used in object detection task because of its simple and effective structure. Resnet introduces the design of deep residual block to connect the bottom feature jump to the top feature, which overcomes the problem that the gradient disappears and the accuracy cannot be effectively improved due to the deepening of network depth. In order to achieve the balance between speed and accuracy, we choose resnet-18 as the backbone. At the same time, we also introduced the asymmetric revolution block (ACBlock) in ACNet[12] published in 2019. We combine ACBlock with resnet-18, as shown in Figure 2. ACBlock uses 3 x 3 convolution blocks, 1 x 3 convolution blocks and 3 x 1 convolution blocks to stack up instead of all 3 x 3 convolution blocks in Resnet. In the training stage, a variety of irregular convolution filters are used for the feature map, which makes the extracted features more diverse and richer, and improves the feature expression ability in encoder stage. Although some additional parameters are added and the training time is lengthened, the irregular convolution kernels of the same layer can be added directly in the reasoning stage without any computational complexity.

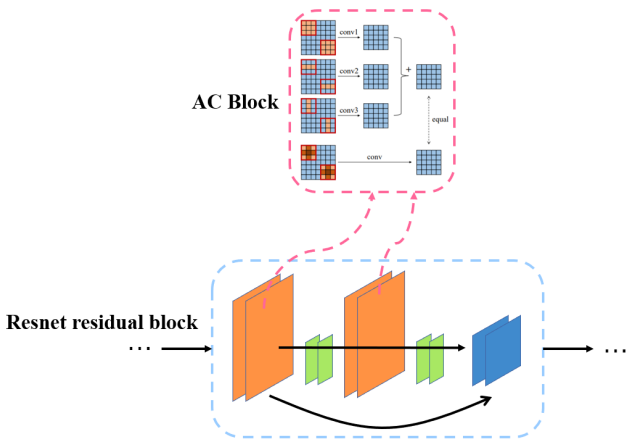

**Figure 2. ACBlock combined with resnet-18**

## 2.3 Feature Pyramid Network

Because the object of remote sensing data set has a large size span, we need to learn image features from feature pyramid network(FPN)[13]. Therefore, we introduced FPN, which is combined with Resnet. Most of the original object detection algorithms only use the top-level features for prediction, but we know that the feature semantic information of the lower level is relatively small, but the target location is accurate; the feature semantic information of the higher level is relatively rich, but the target location is relatively rough. FPN can improve the accuracy of the model by combining the low-level features with the high-level features.

Since the input image is extracted from different scales by FPN, how to use these feature maps is particularly important. There are two common processing methods. The first one is to predict the features of each layer extracted by FPN, and then fuse the features of each layer. The second is to fuse the features of each layer to form the final feature map, and predict the final result of this feature map. In the field of object detection, most of the anchor based models adopt the former method, as shown in Figure 3 (a). In the RPN stage, the regression coefficient and confidence degree of each anchor are predicted for each layer of feature map. The regression coefficient is applied to the coordinates of the initial anchor to obtain the prediction coordinates of all the anchors. Then, NMS algorithm is used to filter out the good bounding boxes from all the anchors, and then the following steps are carried out. In this way, we can filter enough bounding boxes from a large number of multi-level prediction bounding boxes, and reduce the time cost. However, the model of anchor free is different from that based on anchor. The model of anchor free based on center point detection is a direct prediction of center point probability graph, center point offset graph and length width dimension graph. This method can not directly predict the boundary box coordinates, and it is not convenient to stack the prediction results of all feature layers and then filter them. Therefore, we adopt the latter method, as shown in Figure 3 (b). The final feature map is obtained by using 1 x 1 convolution to reduce the number of channels.

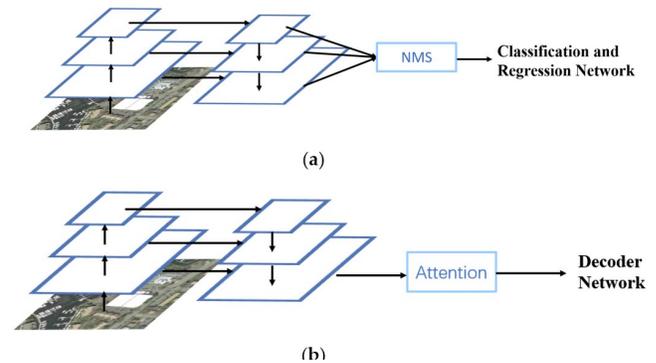

**Figure 3. There are two ways to use feature pyramids. (a) is often used in the model based on anchor. (b) is used in CenterFPANet.**

## 2.4 Channel Attention

Although the convolutional neural network has strong non-linear expression ability, when the amount of information is too complex and more information needs to be processed, the corresponding model also needs to be deeper to obtain stronger expression ability. In order to reduce the complexity of the model, we can use attention mechanism to improve the ability of neural network to process information by the way of human brain processing overload information. Attention mechanism can make the model focus on more important information and ignore the unimportant information. In fact, the realization of attention mechanism is divided into three steps: information input, calculation of attention distribution, and calculation of weighted average of input information based on attention distribution. The general expression of attention mechanism can be as follows:

$$O = f(QK^T)V \quad (1)$$

$Q$ is the query term matrix, $K$ is the corresponding key term, $V$ is the value term, and $f$ is the activation function. Attention mechanism can be universally understood as a layer of perceptron composed of $Q$, $K$ and $V$, and its activation function is $f$.

We hope that the attention used can give appropriate weight to the multi-level feature channel extracted by FPN, so that the model can dynamically adjust the weight relationship between the high-level feature and the low-level feature according to the features of the input image. Here, the key to the problem of $K = V$ = feature map is how to construct $Q$. We choose SE block[14]ff as $Q$ to construct the query matrix. SE block is mainly composed of two parts, namely 'squeeze' and 'excitation'. The nonlinear calculation of these two parts to the original image can be seen as the product of the original feature map $K$ and SE block. Then, through sigmoid for weight reduction, and then multiply with the original feature image $V$, the new feature maps after weight change is obtained.

In the squeeze step, Se block compresses the global spatial information into a channel descriptor to increase the receptive field. Compression is done by global average pooling. The next execution operation is used to activate the information in the aggregate feature. Se block adopts a simple gathering mechanism with a sigmoid activation, which can learn the non-linear relationship between channels and non-mutually exclusive relationship, so as to ensure that multiple channels can act simultaneously.

The detailed implementation algorithm is as follows.

SE block can be embedded in all kinds of models flexibly. The author of the original paper recommends that it should be embedded in the residual structure of Resnet. Considering the efficiency of our model, we connect directly behind the FPN structure. In the model, the multi-layer feature map predicted by FPN is directly sampled and concatenate to a uniform size, and then input to SE block to get the final feature map. Then, after 1 x 1 convolution to reduce the channel dimension, heatmaps, width and height graphs and center-regression graphs are predicted directly. The detailed SE block structure is shown in Figure 4.

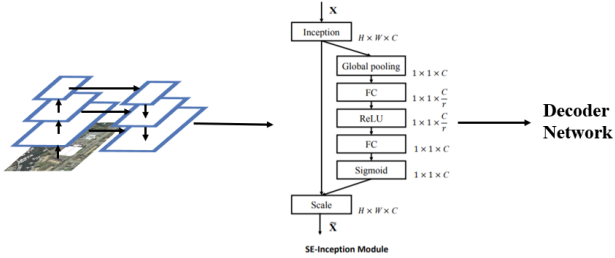

Figure 4. Channel attention mechanism combined with backbone.

## 2.5 Inference

In the inference stage, the final bounding box category, bounding box coordinates and corresponding confidence need to be obtained from the heatmap and wh(width and height), offset reg feature maps predicted by the model. The category and confidence of the object are derived from the heatmap. If the value of a pixel on the heatmap is larger than the surrounding eight pixels, then this pixel is called a hot spot. In the $W * H * C$ heatmap output by the model, each channel can extract several hot spots. We perform a $3\times3$ max pooling on the heatmap to obtain all hot spots of the entire image, and take the largest K values as object center points (K is the hyperparameter), denoted as $(\widehat{x_c}, \widehat{y_c})$. The corresponding channel is the category of the object, and the corresponding value is the confidence of the object. After obtaining hotspots in each channel, the width, height, and pixel offset of the corresponding position can be directly obtained in the tensor of the $W * H * (2 + 2)$ dimension output by object size (height and width) regression branch and the offset regression branch, denoted as $(\widehat{w_c}, \widehat{h_c})$ and $(\widehat{w_{off}}, \widehat{h_{off}})$. So the object position $(\widehat{x_1}, \widehat{y_1}, \widehat{x_2}, \widehat{y_2})$ detected by the model can be calculated by the following formula:

$$\widehat{x_1} = \widehat{x_c} + \widehat{w_{off}} - \frac{\widehat{w_c}}{2} \quad (2)$$

$$\widehat{x_2} = \widehat{x_c} + \widehat{w_{off}} + \frac{\widehat{w_c}}{2} \quad (3)$$

$$\widehat{y_1} = \widehat{y_c} + \widehat{h_{off}} - \frac{\widehat{h_c}}{2} \quad (4)$$

$$\widehat{y_2} = \widehat{y_c} + \widehat{h_{off}} + \frac{\widehat{h_c}}{2} \quad (5)$$

Finally, up-sampling the above coordinates back to the original size to complete the output of the detected objects, the flow of the inference stage is shown in the figure below.

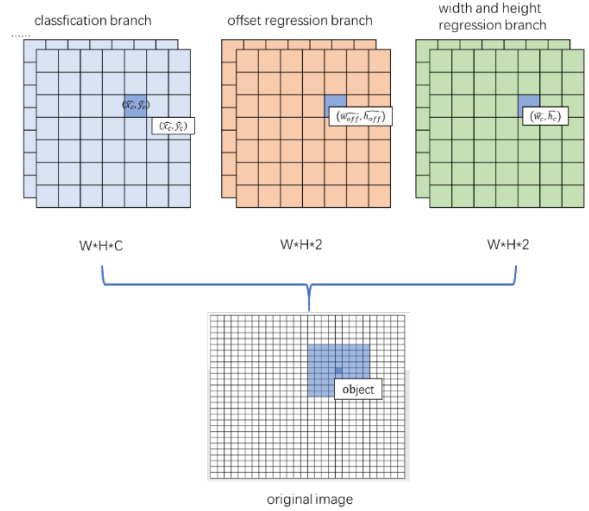

Figure 5. Output of decoder network.

## 2.6 Losses

The output of the model is a tensor of dimension $W * H * C + W * H * (2 + 2)$, where W, H, and C are the width, height, and the number of object categories, respectively. The first '2' represents the height and width of the object centered on each position of the feature map, and the second '2' represents the offset of the horizontal and vertical coordinates of the corresponding pixel position. Due to down sampling, the pixel value of the object center point after down sampling only retains the integer part, and the detection of small objects will affect the accuracy, so this offset is added to correct the error of integer truncation. In order to deal with the problem of imbalance between positive and negative samples, this paper adopts an improved focal loss[15] as loss function for classification branch; for the regression training of height, width and offset, this paper adopts L1 loss; the final loss is the sum of the two.

For a common object detection dataset, the ground truth format is mostly: $x1, y1, x2, y2, \text{label}$. The coordinates of two points are used to describe the object position, but the output of this model is a $W * H * (C + 2 + 2)$ dimension tensor, so before calculating the loss, it is needed to convert the ground truth into the consistent form firstly. When calculating the ground truth of $W * H * C$ dimension tensor, calculating the center point coordinates according to the original ground truth(Formula 6), then down-sampling R times (here $R = 4$), the center point pixel of the low-resolution feature map after downsampling is recorded as q(Formula 7), then the ground truth in heat maps corresponding to q pixels all equal to 1, and the rest of the pixels should be 0 because of the non-center point for any objects. But in fact, even if the model outputs is a few pixels nearby, the result of object detection is still very accurate, this design would make training difficult. Therefore, after setting the hot spots corresponding to the q pixel to 1, the Gaussian kernel diffusion method is used to diffuse the ground truth value to the entire $W * H * C$ heatmap(Formula 8), where $\delta_1, \delta_2$ are the standard deviations related to the object size. If in one heatmap, multiple Gaussian distributions overlap, the larger one is directly taken. Through this conversion, in the heatmap, the position at the object's center point is 1, and the value of the surrounding pixels gradually decrease to 0. When calculating the ground truth corresponding to height and width ($W * H * 2$ tensor), simply record the object height and width after down sampling at the corresponding position. When calculating the ground truth corresponding to offset ($W * H * 2$ tensor), just assigning the decimal part of the horizontal and vertical coordinates after down sampling to the corresponding position This completes the ground truth conversion.

$$p = \left(\frac{x1 + x2}{2}, \frac{y1 + y2}{2}\right) \quad (6)$$

$$q = \left\lfloor \frac{p}{R} \right\rfloor \quad (7)$$

$$Y_{xyc} = \exp\left(-\frac{(x - q_x)^2}{2\delta_1^2}\right) * \exp\left(-\frac{(y - q_y)^2}{2\delta_2^2}\right) \quad (8)$$

Among them, $x1, y1, x2, y2$ in formula (6) are the pixel coordinates of the top left corner and bottom right corner of the object, and p is the coordinate of the object center point. R in formula (7) is the down sampling scale, and q is the coordinate of the corresponding pixel after down sampling R times at the center point p. $Y_{xyc}$ in formula (8) represents the ground truth of the pixel of the Cth channel whose coordinates are (x, y), the range is [0,1], $q_x, q_y$ represents the coordinate of the center point after down sampling, the value of the pixel is the maximum value 1, $\delta_1, \delta_2$ are calculated by the height and width of the object. To ensure that the bounding boxes output by the model has a large IOU with the actual object, let $3\delta_1 = \frac{W'}{2}, 3\delta_2 = \frac{H'}{2}$ in this article, where $W', H'$ are the width and height of the object after down sampling. Most of the ground truth after Gaussian kernel diffusion can fall in the central area of the object.

### 2.6.1 Heatmap Loss（Classification Loss Function）

For the calculation of classification loss, the output of the classification branch of this model is the probability that each pixel is the object center point. The model will regard the pixel points whose ground truth exceeds the threshold value $\varphi$ in the heatmap (that is, the corresponding position of the object center point after down sampling) as positive samples, all other pixels with ground truth less than $\varphi$ are regarded as negative samples, so the number of positive and negative samples is extremely uneven. In order to solve this problem, this paper uses improved focal loss as the loss function of classification branch, as shown in the following formula.

$$L_c = -\frac{1}{N}\sum \begin{cases} (1 - \widehat{Y_{xyc}})^\alpha \log(\widehat{Y_{xyc}}), Y_{xyc} \geq \varphi \\ (1 - Y_{xyc})^\beta (\widehat{Y_{xyc}})^\alpha \log(1 - \widehat{Y_{xyc}}), Y_{xyc} < \varphi \end{cases} \quad (9)$$

The original focal loss is suitable for binary classification, that is, the ground truth of positive samples is $y = 1$, and the ground truth of negative samples is $y = 0$. However, the ground truth of positive and negative samples in this model is an interval ($Y \geq \varphi$ and $Y < \varphi$), Therefore, the value of the ground truth is involved in the calculation of the loss function, so that the weight of the pixel points that are far from the center point and the prediction probability output by the model is large can be set higher, and the loss of such negative samples is increased, which is conducive to the loss function optimization process. Generally, the threshold $\varphi$ can be between 0.8 and 1 according to the actual application. The larger $\varphi$, the more positive samples, but the less accurate in model training. This model considers that the heatmap is obtained by down sampling R ($R = 4$) times from the original image, and there is a certain deviation itself. If $\varphi$ is small, the model is difficult to extract a more accurate center point, so just let $\varphi = 1$, That is, pixels with a ground truth equals 1 in the original heatmap are regarded as positive samples, and the rest are regarded as negative samples.

### 2.6.2 Width-height Loss and Offset Loss

For the regression loss calculation of height, width and offset, because these values are relatively large, so we use L1 loss. As shown in the formula below.

$$L_{wh} = \frac{1}{N}\sum |Y_{wh} - \widehat{Y_{wh}}| \quad (10)$$

$$L_{off} = \frac{1}{N}\sum |Y_{off} - \widehat{Y_{off}}| \quad (11)$$

### 2.6.3 Total Loss

The overall objective function of the model is the weighted sum of $L_c$, $L_{wh}$ and $L_{off}$. Considering that the classification branch loss and offset branch loss values are small, and the height and width branch loss values are relatively large, so set different weights, this model sets $\lambda_1 = 0.1$, $\lambda_2 = 1$.

$$L = L_c + \lambda_1 L_{wh} + \lambda_2 L_{off} \quad (12)$$

## 3. Results

This section reports several experiments conducted with the proposed architecture.

### 3.1 Dataset

In order to verify the effect of the model, we tested it on the DOTA dataset[16]. Dota is a large-scale dataset for object detection in aerial images. The images of in DOTA-v1.0 dataset are manily collected from the Google Earth, some are taken by satellite JL-1, the others are taken by satellite GF-2 of the China Centre for Resources Satellite Data and Application. It contains 2806 aerial images, the size of each image is in the range of about 800 × 800 to 4000 × 4000 pixels, and contains objects of various proportions, directions and shapes. There are 15 different categories in the dataset. The object categories in DOTA-v1.0 include: plane, ship, storage tank, baseball diamond, tennis court, basketball court, ground track field, harbor, bridge, large vehicle, small vehicle, helicopter, roundabout, soccer ball field and swimming pool. The categories of the dataset are shown in Figure 6(**a**). It can be seen that in all samples, more than 80% are vehicles, cars, ships and storage-tank, and the remaining 11 categories have a small number of samples. This caused difficulties in training and testing. Moreover, as shown in Figure 6(**b**), the sample size scales of different categories and the same category vary greatly, and it is difficult to choose an appropriate segmentation size to segment the original input image. Therefore, we have adopted multi-scale detection to deal with this problem. Due to the limitation of computing resources, we chose to split the input image into a size of 1024 * 1024 and use 824 as the stride. During prediction, the input image is scaled at different scales to reduce the difficulty of prediction caused by the different sample sizes.

The entire splitted dataset is divided into two parts, namely training set and validation set. The separation method is to randomly sample 80% of the data as training set and 20% as validation set. For the training set, in order to further expand the dataset scale, we used color enhancement and random cropping for data augmentation. For the validation set, we adopt the same image cutting method, the size is 1024 * 1024, and the stride is 824. Finally, the splitted prediction images are merged into the entire image, and then NMS is used to remove the overlapping bounding boxes at the borders of two adjacent images.

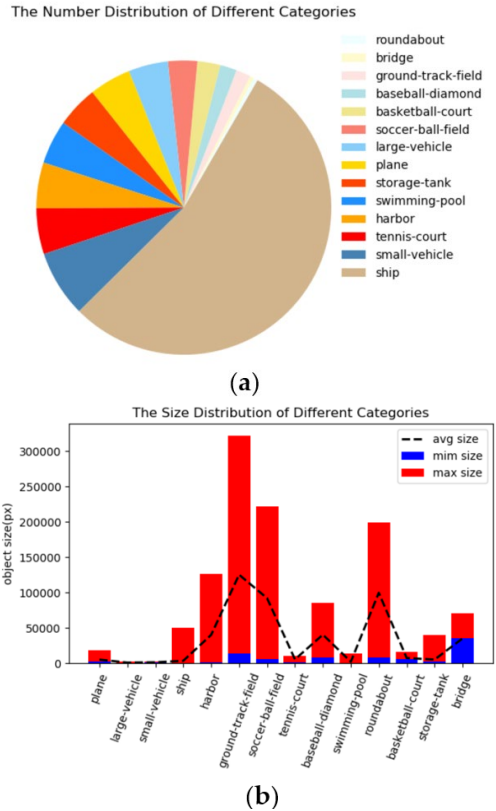

**Figure 6. (a) The number distribution of objects in different categories; (b) the size distribution of objects in different categories.**

## 3.2 Experiment Environment
The experiments in this paper were conducted on E5-2650, 132G memory, 4 TITAN XP workstations.

## 3.3 Evaluation Method
The evaluation indicators used in the experiments are mAP, speed, model parameters, etc.

mAP is a commonly used evaluation index in the field of object detection. mAP chooses the average value of precision under 11 different recall values ([0, 0.1, ..., 0.9, 1.0]). Precision is the proportion of positive samples among all predicted samples. Recall is the proportion of all positive samples that are detected. The method to judge whether the predicted sample is a positive sample is to calculate the IOU of the predicted bounding box and the true bounding box. If the IOU is greater than the artificially set threshold, the predicted sample is a positive sample. IOU is an index to measure the overlapping degree of two bounding boxes.

$$AP = \frac{1}{11} \sum_{r \epsilon \{0, 0.1, \dots 1\}} p_{interp}(r) \quad (13)$$

$$precision = \frac{TP}{TP + FP} \quad (14)$$

$$recall = \frac{TP}{TP + FN} \quad (15)$$

Speed is used to measure the detection speed of the model. The index to evaluate the speed of the model is the time taken (MS) to predict each picture on average.

The amount of model parameters can measure the lightness of the model. The small amount of model parameters means that the model is more efficient and requires less computing resources.

## 3.4 Evaluation Results
The evaluation results of all the experiments in this section are performed on the DOTA valid set. The evaluation result (mAP) on the valid set is calculated using the official development kit provided by DOTA. The backbone of each experimental model is loaded with pre-trained weight files, which is pretrained by ImageNet dataset. The experimental results of our model always use the results of the 180th generation. The initial value of learning rate is 1.25e-4, and the optimizer is Adam optimizer, the maximum number of objects per image is set to 160. The model is detected on the 1024 * 1024 small image after cutting. When merging into the original image, NMS is needed. Here, the IOU threshold of NMS is set to 0.45.

We evaluate the improvement of each module of the model separately. The baseline is the CenterNet that introduced ACBlock. Table 1 shows the mAP and AP of each category, including baseline model, the model with FPN module and the model with FPA module(combined FPN and Attention module).

The baseline model architecture is CenterNet with the asymmetric revolution block(ACBlock). The 3 × 3 convolution kernel in the original Resnet network is replaced by 3 × 1, 1 × 3, 3 × 3 convolution kernels, to extract more diverse feature maps, and then connect the classification branch and regression branch to get the final detection result. From Tabel 1, mAP is 62.26%, the detection accuracy in some categories, such as planet, large vehicle, tennis court, storage tank, etc. performs good.

Table 1. Accuracy performance of different models

| Method | Baseline | Baseline+FPN | Baseline+FPA | Baseline+FPA+mutiscaletest |
|---|---|---|---|---|
| Plane | 88.90 | 89.14 | 88.83 | 88.74 |
| BD | 66.83 | 69.02 | 69.63 | 71.25 |
| Bridge | 43.42 | 45.18 | 47.22 | 48.95 |
| GTF | 50.19 | 50.88 | 45.80 | 52.06 |
| SV | 49.14 | 50.59 | 49.96 | 48.55 |
| LV | 70.07 | 73.62 | 73.89 | 73.37 |
| Ship | 65.09 | 66.44 | 63.46 | 61.14 |
| TC | 90.52 | 90.48 | 90.44 | 90.53 |
| BC | 61.92 | 54.80 | 59.52 | 57.63 |
| ST | 75.29 | 82.15 | 83.65 | 84.06 |
| SBF | 58.75 | 57.06 | 58.40 | 66.64 |
| RA | 61.52 | 64.38 | 64.25 | 62.71 |
| Harbor | 65.98 | 65.13 | 69.43 | 73.33 |
| SP | 46.67 | 50.16 | 55.00 | 57.63 |
| Helicopter | 39.63 | 46.07 | 40.55 | 42.76 |
| mAP | 62.26 | 63.67 | 64.00 | 65.29 |

After adding FPN on the baseline, that is, after Resnet-18 the feature pyramid network is used to obtain the features of different levels, and then deconvolution to the same size and concatenate them, and then connect the classification and regression branches after multi-layer feature maps to obtain the test results. Compared with baseline, after obtaining multi-layer feature maps, features of different scales can be extracted. Storage tank, helicopter and other categories have a good improvement, and the overall mAP is increased to 63.67%.

Further, after introducing the attention mechanism on the basis of FPN, the weights of different channels are dynamically adjusted according to the characteristics of the input remote sensing image, and the most useful features are sent to the classification and regression branches. Similarly, training 180 epoch, storage tank, swimming pool and other categories improved significantly, and mAP reached 64.00%.

Analysis shows that after adding FPA module (FPN + attention) to the baseline, it can obtain higher accuracy than the baseline, which verifies the effectiveness of the FPA module. Using multi-level detection in the inference stage, mAP has further increased to 65.29%. Compared with the baseline, most categories of AP have more improvements. However, due to the large number of objects such as Plane, small vehicles, and large vehicles, or the original AP has reached a high level, these categories have not been significantly improved.

We compared the inference speed, the amount of model parameters and mAP of the CenterFPANet with other common models. For YOLOv2 and YOLOv3, first we use K-means clustering algorithm to calculate the initial size of the anchors, and then train. The training parameters basically refer to the official Darknet[17], the initial learning rate is set to 0.001, iterative to the loss function convergence. We selected the 400,000th epoch's model with the highest mAP as the experimental result. The models of RetinaNet, Faster RCNN, and MaskRCNN are provided by Mmdetection[18]. The relevant parameters of training also refer to official parameters. The initial value of learning rate is 0.01, which is decayed in the 16th and 22nd epochs respectively. Finally, the 24th epoch model with loss function convergence is selected for evaluation. The experimental results are shown in Table 2.

Table 2. Time consumption of different models

| Method | Speed(fps) | Size(MB) | mAP |
|---|---|---|---|
| YOLOv2 | 20.2 | 240 | 42.72 |
| YOLOv3 | 9.0 | 262 | 37.42 |
| Mask RCNN | 8.3 | 391 | 71.61 |
| Faster RCNN | 13.8 | 380 | 70.76 |
| RetinaNet | 12.2 | 367 | 67.45 |
| CenterFPANet | 22.2 | 113 | 64.00 |

YOLOv2 and YOLOv3 are both classic one-stage object detectors, which divide the entire image evenly into multiple grids, and directly predict N anchor sizes for each grid. Therefore, the detection speeds are 20.2fps and 9.0fps, respectively. However, due to the large scale of object size and the denseness of the objects in DOTA dataset, the YOLO series network is easy to miss small objects and dense objects, so it is not suitable for remote sensing images. They only reached 42.72% and 37.42% mAP.

Faster RCNN and RetinaNet are classic anchor-based object detectors, which need to extract candidate anchors before making predictions. Based on Faster RCNN, RetinaNet introduces focal loss to solve the problem of sample imbalance. But in this experiment, the effect of Faster RCNN is better, reached 70.76%, which is 3.31% higher than RetinaNet. It shows that focal loss cannot solve the problem of sharp imbalance in the number of various samples in the DOTA dataset. In terms of speed, the anchor-based detectors are the slowest, only 13.8fps and 12.2fps. Mask RCNN is a semantic segmentation network based on Faster RCNN. Therefore, it is more accurate than Faster RCNN, but the speed is slowed by 5.5fps.

CenterFPANet is also a one-stage object detection network, but compared to the YOLO series, our network is an anchor free detector, which directly predicts the location of the object center points, so the speed can be as high as 22.2fps, which is 1.9fps faster than YOLOv2. Compared with the two-stage network, it is almost twice as fast. In terms of accuracy, our network is more suitable for detecting dense objects and multi-scale objects, which is 21.3% higher than YOLOv2. CenterFPANet's mAP is 64.00%, which is 7.61% lower than Mask RCNN, but its speed is 2.67 times faster. Therefore, it can be said that CenterFPANet has reached a balance between speed and accuracy.

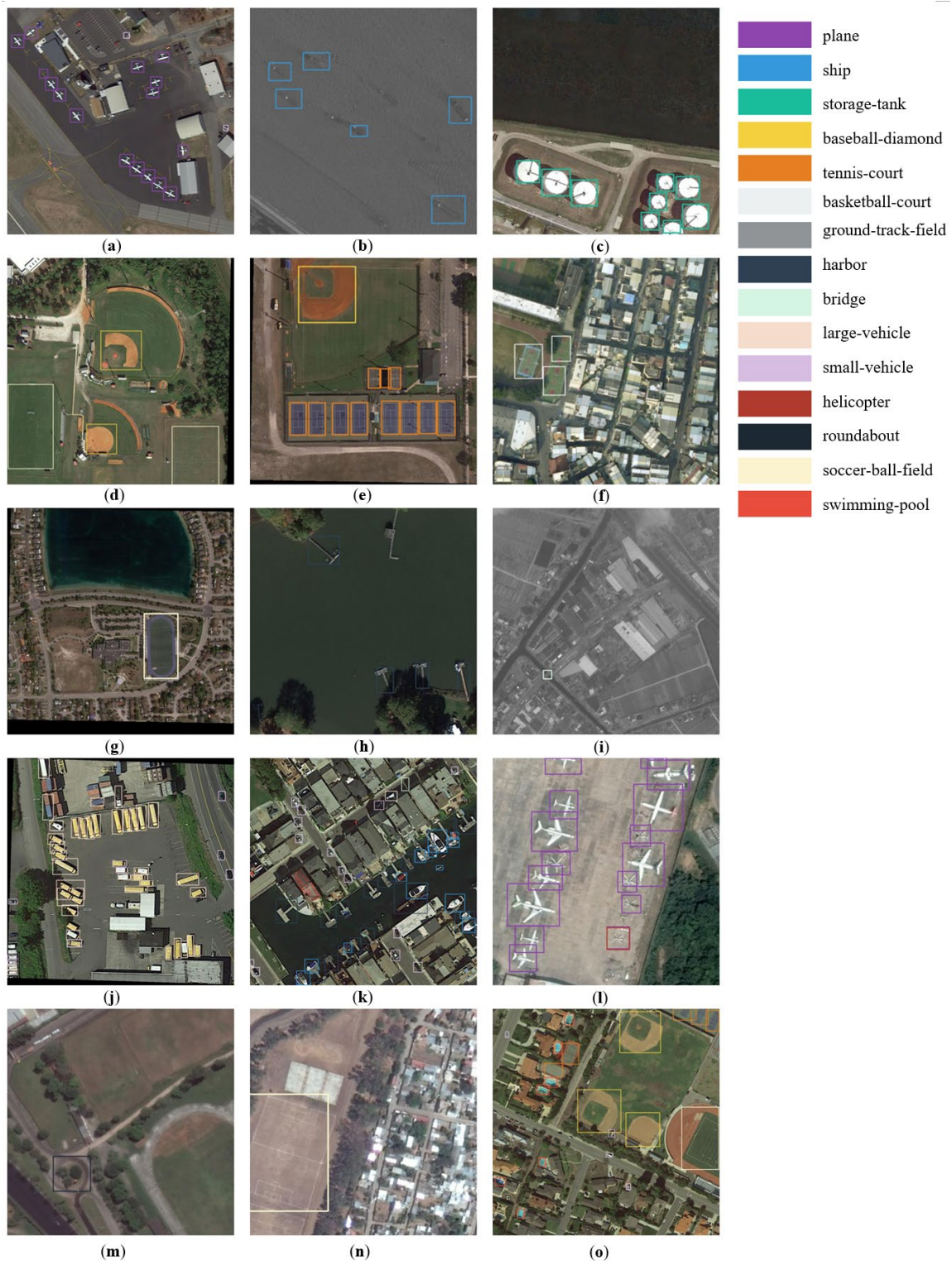

**Figure 7. Visualization of detection results from CenterFPANet**

In terms of the size of the model, our network also has outstanding advantages. Our model parameters are only 113MB, only half of the parameters of the YOLO series, and one third of the RCNN series.

We visualized the partial results of CenterFPANet in the DOTA dataset, as shown in Figure x. Since some object sizes are too small, we appropriately cropped 1024 size images. Each picture shows the predicted bounding boxes for each category.

The typical characteristics of remote sensing images can be seen from above figure. The sharpness of the image varies, and the sharpness of Figure 7(i) is reduced due to nighttime shooting and blocked by clouds. The size of objects in different categories varies greatly, and the size of plane in Figure 7(a), and the baseball diamond and soccer ball field in Figure 7(d) and Figure 7(n) are quite different. The objects are relatively dense, and the objects in Figure 7(j) are awfully close, which may cause missed detection. Some categories are relatively similar in shape, such as basketball court and soccer ball field, it is difficult to distinguish.

Faced with the unique difficulties of these remote sensing images, CenterFPANet solved these problems well. For scenes with dense objects, as shown in Figure 7(j) and Figure 7(l), we can see that our model has less missing detection problems. In a multi-classification scenario, as shown in Figure 7(k), the task can also be completed well. Of course, CenterFPANet also has some problems. As shown in Figure 7(h), the model missed the harbor in the upper right corner. We can observe that the shapes of these harbors are quite different, which makes detection difficult. This shows that CenterFPANet's generalization ability is not extraordinarily strong.

## 4. Discussion

We have proved the effectiveness of the FPA module through experiments. FPN enables the network to detect large and small objects more accurately at the same time. In the anchor free detection method, FPN is used to extract the characteristics of remote sensing images. It is impossible to directly generate anchors artificially at various level features, like anchor-based methods such as Faster RCNN. Given this problem, we introduced an attention mechanism, which is based on the characteristics of different images, adjust the weight of each channel dynamically to extract more useful features for subsequent detection subnetwork.

Compared with other models, our speed and model parameters have outstanding advantages. The model capacity of only 110MB can be more easily embedded in the drone, and the speed of 22.20fps can ensure real-time detection. At the same time, CenterFPANet can still maintain a high accuracy rate.

In summary, compared with other models, CetnerFPANet is undoubtedly the most suitable object detection network for detecting remote sensing images.

## 5. Conclusions

In this paper, we discuss the problem of slow detection speed of the existing optical remote sensing image object detection model. To solve this problem, we propose a multi-classification object detection framework for optical remote sensing images. We adopt the strategy of lightweight encoder and multi-scale attention mechanism. With this strategy, the accuracy of network detection can be enhanced without huge loss of detection speed. Experiments show that the proposed framework is efficient and accurate for multi-scale objects in complex optical remote sensing scenes.

# Columns on Last Page Should Be Made As Close As Possible to Equal Length

## Authors' background

| Your Name | Title* | Research Field | Personal website |
|---|---|---|---|
| Xi Gu | master student | Computer vision | |
| Lingbin Kong | master student | Computer vision | |
| Zhicheng Wang | full professor | Computer vision | https://see.tongji.edu.cn/info/1153/5486.htm |
| Jie Li | senior lecture | Computer vision | |
| Zhaohui Yu | master student | Computer vision | |
| Gang Wei | full professor | Computer vision | |

*This form helps us to understand your paper better, **the form itself will not be published.**

*Title can be chosen from: master student, Phd candidate, assistant professor, lecture, senior lecture, associate professor, full professor